\documentclass{article}
\pdfoutput=1
\usepackage[letterpaper,margin=1in,rmargin=1in,lmargin=1in]{geometry}
\usepackage[utf8]{inputenc}

\usepackage{kpfonts}
\usepackage[T1]{fontenc}
\usepackage{authblk}

\usepackage{hyperref}
\usepackage{url}

\usepackage{amsmath}

\DeclareMathOperator*{\argmin}{argmin}

\usepackage{amssymb}
\usepackage{xspace}
\usepackage{graphicx}
\usepackage{color}
\usepackage{bm}
\usepackage{bbm}
\usepackage[caption=false]{subfig}
\usepackage{wrapfig}
\usepackage{booktabs}

\newcommand{\mcD}{\mathcal{D}}
\newcommand{\bfw}{\mathbf{w}}
\newcommand{\bfx}{\mathbf{x}}
\newcommand{\bbR}{\mathbb{R}}
\newcommand{\bdelta}{\bm{\delta}}

\usepackage{algorithm}
\usepackage{algpseudocode}

 \newcommand{\anew}[1]{{#1}}

\title{Analyzing Federated Learning through an Adversarial Lens}
\author[1]{Arjun Nitin Bhagoji\thanks{Contact: abhagoji@princeton.edu; Work done while at I.B.M. T.J. Watson Research Center}}
\author[2]{Supriyo Chakraborty}
\author[1]{Prateek Mittal}
\author[2]{Seraphin Calo}

\affil[1]{Department of Electrical Engineering, Princeton University}
\affil[2]{I.B.M T.J. Watson Research Center}

\date{}

\begin{document}

\newcommand{\fedlearn}{federated learning\xspace}
\newcommand{\crowdlearn}{crowdsourced learning\xspace}
\maketitle

\begin{abstract}
Federated learning distributes model training among a multitude of agents, who, guided by privacy concerns, perform training using their local data but share only model parameter updates, for iterative aggregation at the server. In this work, we explore the threat of \emph{model poisoning} attacks on federated learning initiated by a single, non-colluding malicious agent where the adversarial objective is to cause the model to mis-classify a set of chosen inputs with high confidence. We explore a number of strategies to carry out this attack, starting with simple \emph{boosting} of the malicious agent's update to overcome the effects of other agents' updates. To increase attack stealth, we propose an alternating minimization strategy, which alternately optimizes for the training loss and the adversarial objective. We follow up by using parameter estimation for the benign agents' updates to improve on attack success. Finally, we use a suite of interpretability techniques to generate visual explanations of model decisions for both benign and malicious models, and show that the explanations are nearly visually indistinguishable. Our results indicate that even a highly constrained adversary can carry out model poisoning attacks while simultaneously maintaining stealth, thus highlighting the vulnerability of the federated learning setting and the need to develop effective defense strategies.
\end{abstract}


\section{Introduction} 
\label{sec: intro}
Federated learning \cite{pmlr-v54-mcmahan17a} has recently emerged as a popular implementation of distributed stochastic optimization for large-scale deep neural network training. It is formulated as a multi-round strategy in which the training of a neural network model is distributed between multiple agents. In each round, a random subset of agents, with local data and computational resources, is selected for training. The selected agents perform model training and share only the parameter updates with a centralized parameter server, that facilitates aggregation of the updates. Motivated by privacy concerns, the server is designed to have no visibility into an agents' local data and training process. The aggregation algorithm used is usually weighted averaging.

In this work, we exploit this lack of transparency in the agent updates, and explore the possibility of an adversary controlling a small number of malicious agents (usually just 1) performing a \emph{model poisoning attack}. The adversary's objective is to cause the jointly trained global model to misclassify a set of chosen inputs with high confidence, i.e., it seeks to poison the global model in a targeted manner. Since the attack is targeted, the adversary also attempts to ensure that the global model converges to a point with good performance on the test or validation data We note that these inputs are not modified to induce misclassification as in the phenomenon of adversarial examples \cite{carlini2017towards,szegedy2013intriguing}. Rather, their misclassification is a product of the adversarial manipulations of the training process. We focus on an adversary which directly performs model poisoning instead of data poisoning \cite{biggio2012poisoning,rubinstein2009stealthy,poison-mt,poison-lasso,poison-mt,koh2017understanding, chen17,jagielski2018manipulating} as the agents' data is never shared with the server. In fact, model poisoning subsumes dirty-label data poisoning in the federated learning setting (see Section \ref{subsec: data_poison} for a detailed quantitative comparison). 

Model poisoning also has a connection to a line of work on defending against Byzantine adversaries which consider a threat model where the malicious agents can send arbitrary gradient updates \cite{peva17,yudong17,mhamdi2018hidden,chen18,yin2018byzantine} to the server. However, the adversarial goal in these cases is to ensure a distributed implementation of the Stochastic Gradient Descent (SGD) algorithm converges to `sub-optimal to utterly ineffective models'\cite{mhamdi2018hidden} while the aim of the defenses is to ensure convergence. On the other hand, we consider adversaries aiming to only cause targeted poisoning. In fact, we show that targeted model poisoning is effective even with the use of Byzantine resilient aggregation mechanisms in Section \ref{sec: byzantine_attack}. Concurrent and independent work \cite{bagdasaryan2018backdoor} considers both single and multiple agents performing poisoning via model replacement, which boosts the entire update from a malicious agent, while we consider adversaries that only boost the malicious component of the update.


\subsection{Contributions}
We design attacks on federated learning that ensure targeted poisoning of the global model while ensuring convergence. Our realistic threat model considers adversaries which only control a small number of malicious agents (usually 1) and have no visibility into the updates that will be provided by the other agents. All of our experiments are on deep neural networks trained on the Fashion-MNIST \cite{xiao2017fashion} and Adult Census\footnote{\url{https://archive.ics.uci.edu/ml/datasets/adult}} datasets. 

\noindent \textbf{Targeted model poisoning:} In each round, the malicious agent generates its update by optimizing for a malicious objective designed to cause targeted misclassification. However, the presence of a multitude of other agents which are simultaneously providing updates makes this challenging. We thus use \emph{explicit boosting} of the malicious agent's which is designed to negate the combined effect of the benign agents. Our evaluation demonstrates that this attack enables an adversary \emph{controlling a single malicious agent} to achieve targeted misclassification at the global model with 100\% confidence while ensuring convergence of the global model for deep neural networks trained on both datasets.

\noindent \textbf{Stealthy model poisoning:} We introduce notions of stealth for the adversary based on accuracy checking on the test/validation data and weight update statistics and empirically shoe that targeted model poisoning with explicit boosting can be detected in all rounds with the use of these stealth metrics. Accordingly, we modify the malicious objective to account for these stealth metrics to carry out stealthy model poisoning which allows the malicious weight update to avoid detection for a majority of the rounds. Finally, we propose an \emph{alternating minimization} formulation that accounts for both model poisoning and stealth, and enables the malicious weight update to avoid detection in almost all rounds.

\noindent \textbf{Attacking Byzantine-resilient aggregation}: We investigate the possibility of model poisoning when the server uses Byzantine-resilient aggregation mechanisms such as Krum \cite{peva17} and coordinate-wise median \cite{yin2018byzantine} instead of weighted averaging. We show that targeted model poisoning of deep neural networks with high confidence is effective even with the use of these aggregation mechanisms.

\noindent \textbf{Connections to data poisoning and interpretability:} We show that standard dirty-label data poisoning attacks \cite{chen17} are not effective in the federated learning setting, even when the number of incorrectly labeled examples is on the order of the local training data held by each agent. Finally, we use a suite of interpretability techniques to generate visual explanations of the decisions made by a global model with and without a targeted backdoor. Interestingly, we observe that the explanations are nearly visually indistinguishable, exposing the fragility of these techniques.

\section{Federated Learning and Model Poisoning} 
\label{sec: fd_setup}
In this section, we formulate both the learning paradigm and the threat model that we consider throughout the paper. 
Operating in the federated learning paradigm, where model weights are shared instead of data, gives rise to the \emph{model poisoning} attacks that we investigate.

\subsection{Federated Learning}\label{subsec: fed_learn}
The federated learning setup consists of $K$ agents, each with access to data $\mcD_i$, where $|\mcD_i| = l_i$. The total number of samples is $\sum_i l_i = l$. Each agent keeps its share of the data (referred to as a \emph{shard}) private, i.e. $\mcD_i = \{\bfx^i_1 \cdots \bfx^i_{l_i}\}$ is not shared with the server $S$. The \anew{server is attempting to train a classifier $f$ with} global parameter vector $\bfw_G \in \bbR^n$, where $n$ is the dimensionality of the parameter space. This parameter vector is obtained by distributed training and aggregation over the $K$ agents with the aim of generalizing well over $\mcD_{\text{test}}$, the test data. Federated learning can handle btoh i.i.d. and non-i.i.d partitioning of training data.

At each time step $t$, a random subset of $k$ agents is chosen for synchronous aggregation \cite{pmlr-v54-mcmahan17a}. Every agent $i \in [k]$, \emph{minimizes \footnote{approximately for non-convex loss functions since global minima cannot be guaranteed} the empirical loss over its own data shard} $\mcD_i$, by starting from the global weight vector $\bfw_G^t$ and running an algorithm such as SGD for $E$ epochs with a batch size of $B$. At the end of its run, each agent obtains a local weight vector $\bfw_i^{t+1}$ and computes its local update $\bm{\delta}_i^{t+1}=\bfw_i^{t+1}-\bfw_G^t$, which is sent back to the server. To obtain the global weight vector $\bfw_G^{t+1}$ for the next iteration, any aggregation mechanism can be used. In Section \ref{sec: poison_strategies}, we use weighted averaging based aggregation for our experiments:
$
	\bfw_G^{t+1} = \bfw_G^{t} +  \sum_{i \in [k]} \alpha_i \bm{\delta}_i^{t+1},
$
where $\frac{l_i}{l} = \alpha_i$ and $\sum_i \alpha_i =1$. In Section \ref{sec: byzantine_attack}, we study the effect of our attacks on the Byzantine-resilient aggregation mechanisms `Krum' \cite{peva17} and coordinate-wise median \cite{yin2018byzantine}.

\subsection{Threat Model: Model Poisoning}\label{subsec: model_poison}
Traditional poisoning attacks deal with a malicious agent who poisons some fraction of the \emph{data} in order to ensure that the learned model satisfies some adversarial goal. We consider instead an agent who poisons the \emph{model updates} it sends back to the server. 


\noindent \textbf{Attack Model:} We make the following assumptions regarding the adversary: (i) they control exactly one non-colluding, malicious agent with index $m$ (limited effect of malicious updates on the global model); (ii) the data is distributed among the agents in an i.i.d fashion (making it easier to discriminate between benign and possible malicious updates and harder to achieve attack stealth); (iii) the malicious agent has access to a subset of the training data $\mcD_m$ as well as to auxiliary data $\mcD_{\text{aux}}$ drawn from the same distribution as the training and test data that are part of its adversarial objective. Our aim is to \emph{explore the possibility of a successful model poisoning attack even for a highly constrained adversary}.

\noindent \textbf{Adversarial Goals:} The adversary's goal is to \emph{ensure the targeted misclassification of the auxiliary data} by the classifier learned at the server. The auxiliary data consists of samples $\{\bfx_i\}_{i=1}^r$ with true labels $\{y_i\}_{i=1}^r$ that have to be classified as desired target classes $\{\tau_i\}_{i=1}^r$, implying that the adversarial objective is
\begin{align}\label{eq: adv_obj_comb}
\mathcal{A}(\mcD_m \cup \mcD_{\text{aux}},\bfw^t_G) = \max_{\bfw^t_G} \sum_{i=1}^r \mathbbm{1}[ f(\bfx_i; \bfw^t_G)=\tau_i] .
\end{align}
We note that in contrast to previous threat models considered for Byzantine-resilient learning, the adversary's aim is not to prevent convergence of the global model \cite{yin2018byzantine} or to cause it to converge to a bad minimum \cite{mhamdi2018hidden}. Thus, any attack strategy used by the adversary must \emph{ensure that the global model converges to a point with good performance on the test set}. Going beyond the standard federated learning setting, it is plausible that the server may implement measures to detect aberrant models. To bypass such measures, the adversary must also \emph{conform to notions of stealth} that we define and justify next.

\subsection{Stealth metrics} \label{subsec: stealth}
\anew{Given an update from an agent, there are two critical properties that the server can check. First, the server can verify whether the update, in isolation, would improve or worsen the global model's performance on a validation set. Second, the server can check if that update is very different statistically from other updates. We note that neither of these properties is checked as a part of standard federated learning but we use these to raise the bar for a successful attack.}

\noindent \textbf{Accuracy checking:} The server checks the validation accuracy of $\bfw_i^t=\bfw_G^{t-1}+\bdelta_i^t$, the model obtained by adding the update from agent $i$ to the current state of the global model. If the resulting model has a validation accuracy much lower than that of the model obtained by aggregating all the other updates, $\bfw_{G\setminus i}^t=\bfw_G^{t-1}+\sum_i \bdelta_i^t$, the server can flag the update as being anomalous. For the malicious agent, this implies that it must satisfy the following in order to be chosen at time step $t$:
\begin{align}\label{eqn: accuracy_constraint}
\sum_{\{\bfx_j,y_j\} \in \mcD_{\text{test}}} \mathbbm{1}[f(\bfx_j;\bfw_i^t)=y_i] - \mathbbm{1}[f(\bfx_j;\bfw_{G\setminus i}^t)=y_i] < \gamma_t,
\end{align}
where $\gamma_t$ is a threshold the server defines to reject updates. This threshold determines how much performance variation the server can tolerate and can be varied over time. A large threshold will be less effective at identifying anomalous updates but an overly small one could identify benign updates as anomalous, due to natural variation in the data and training process.

\noindent \textbf{Weight update statistics:} The range of pairwise distances between a particular update and the rest provides an indication of how different that update is from the rest when using an appropriate distance metric $d(\cdot,\cdot)$. In previous work, pairwise distances were used to define `Krum' \cite{peva17} but as we show in Section \ref{sec: byzantine_attack}, its reliance on absolute, instead of relative distance values, makes it vulnerable to our attacks. Thus, we rely on the full range which can be computed for all agent updates and for an agent to be flagged as anomalous, their range of distances must differ from the others by a server defined, time-dependent threshold $\kappa_t$. In particular, for the malicious agent, we compute the range as $R_m= [\min_{i \in [k] \setminus m} d(\bdelta^t_m, \bdelta_i^t), \max_{i \in [k] \setminus m} d(\bdelta^t_m, \bdelta_i^t)]$. Let $R^l_{\text{min},[k\setminus m]}$ and $R^u_{\text{max},[k\setminus m]}$ be the minimum lower bound and maximum upper bound of the distance ranges for the other benign agents among themselves. Then, for the malicious agent to not be flagged as anomalous, we need that
\begin{align}\label{eqn: range_constraint}
\max \{\vert R^u_m - R^l_{\text{min},[k\setminus m]} \vert, \vert R^l_m - R^u_{\text{max},[k\setminus m]} \vert \} < \kappa_t.
\end{align}
This condition ensures that the range of distances for the malicious agent and any other agent is not too different from that for any two benign agents, and also controls the length of $R_m$. We find that it is also instructive to compare the histogram of weight updates for benign and malicious agents, as these can be very different depending on the attack strategy used. These provide a useful qualitative notion of stealth, which can be used to understand attack behavior.

\subsection{Experimental setup}\label{subsec: exp_setup}
We evaluate our attack strategies using two qualitatively different datasets. The first is an image dataset, Fashion-MNIST \cite{xiao2017fashion} which serves as a drop-in replacement for the commonly used MNIST dataset \cite{lecun1998gradient}, which is not representative of modern computer vision tasks. It consists of $28 \times 28$ grayscale images of clothing and footwear items and has $10$ output classes. The training set contains 60,000 data samples while the test/validation set has 10,000 samples. For this dataset, we use a 3-layer Convolutional Neural Network (CNN) with dropout as the model architecture. With centralized training, this model achieves 91.7\% accuracy on the test set. 

The second dataset is the UCI Adult Census dataset\footnote{\url{https://archive.ics.uci.edu/ml/datasets/adult}} which has over 40,000 samples containing information about adults from the 1994 US Census. The classification problem is to determine if the income for a particular individual is greater (class `0') or less (class `1') than $\$50,000$ a year. For this dataset, we use a fully connected neural network achieving 84.8\% accuracy on the test set \cite{fernandez2014we} for the model architecture.

For both datasets, we study the case with the number of agents $K$ set to 10 and 100. When $k=10$, all the agents are chosen at every iteration, while with $K=100$, a tenth of the agents are chosen at random every iteration. We run federated learning till a pre-specified test accuracy (91\% for Fashion MNIST and 84\% for the Adult Census data) is reached or the maximum number of time steps have elapsed (40 for $k=10$ and 50 for $k=100$). In Section \ref{sec: poison_strategies}, for illustrative purposes, we mostly consider the case where the malicious agent aims to mis-classify a single example in a desired target class ($r=1$). For the Fashion-MNIST dataset, the example belongs to class `5' (sandal) with the aim of misclassifying it in class `7' (sneaker) and for the Adult dataset it belongs to class `0' with the aim of misclassifying it in class `1'. We also consider the case with $r=10$ (Appendix)


\section{Strategies for Model Poisoning attacks}
\label{sec: poison_strategies}
In this section, we use the adversarial goals laid out in the previous section to formulate the adversarial optimization problem. We then show how explicit boosting can achieve targeted model poisoning. We further explore attack strategies that add stealth and improve convergence.


\subsection{Adversarial optimization setup}
\anew{From Eq. \ref{eq: adv_obj_comb}, two challenges for the adversary are immediately clear. First, the objective represents a difficult combinatorial optimization problem so we relax Eq. \ref{eq: adv_obj_comb} in terms of the cross-entropy loss for which automatic differentiation can be used. Second, the adversary does not have access to the global parameter vector $\bfw^t_G$ for the current iteration and can only influence it though the weight update $\bdelta^t_m$ it provides to the server $S$. So, it performs the optimization over $\hat{\bfw}_G^{t}$, which is an \emph{estimate} of the value of $\bfw_G^{t}$ based on all the information $\mathcal{I}^{t}_m$ available to the adversary. The objective function for the adversary to achieve targeted model poisoning on the $t^{\text{th}}$ iteration is
\begin{equation}\label{eq: adv_opt}
\begin{aligned}
 \argmin_{\bdelta^{t}_m} \quad & L(\{\bfx_i,\tau_i\}_{i=1}^r, \hat{\bfw}^t_G),  \\ 
\text{s.t.} \quad & \hat{\bfw}_G^{t} = g(\mathcal{I}^{t}_m),
\end{aligned}
\end{equation}
where $g(\cdot)$ is an estimator. For the rest of this section, we use the estimate $\hat{\bfw}_G^{t} = \bfw_G^{t-1} +  \alpha_m \bdelta^{t}_m$, implying that the malicious agent ignores the updates from the other agents but accounts for scaling at aggregation. This assumption is enough to ensure the attack works in practice.}



\begin{figure}[t]
	\centering
	\subfloat[Confidence on malicious objective (poisoning) and the accuracy on validation data (targeted) on the global model]{\resizebox{0.5\textwidth}{!}{\input{latex_plots_new/fmnist_m0_mal_converge_.tex}}}
	\label{subfig: baseline_attack}
	\hspace{0pt}
	\subfloat[Comparison of weight update distributions for benign and malicious agents]{\includegraphics[width=0.4\textwidth]{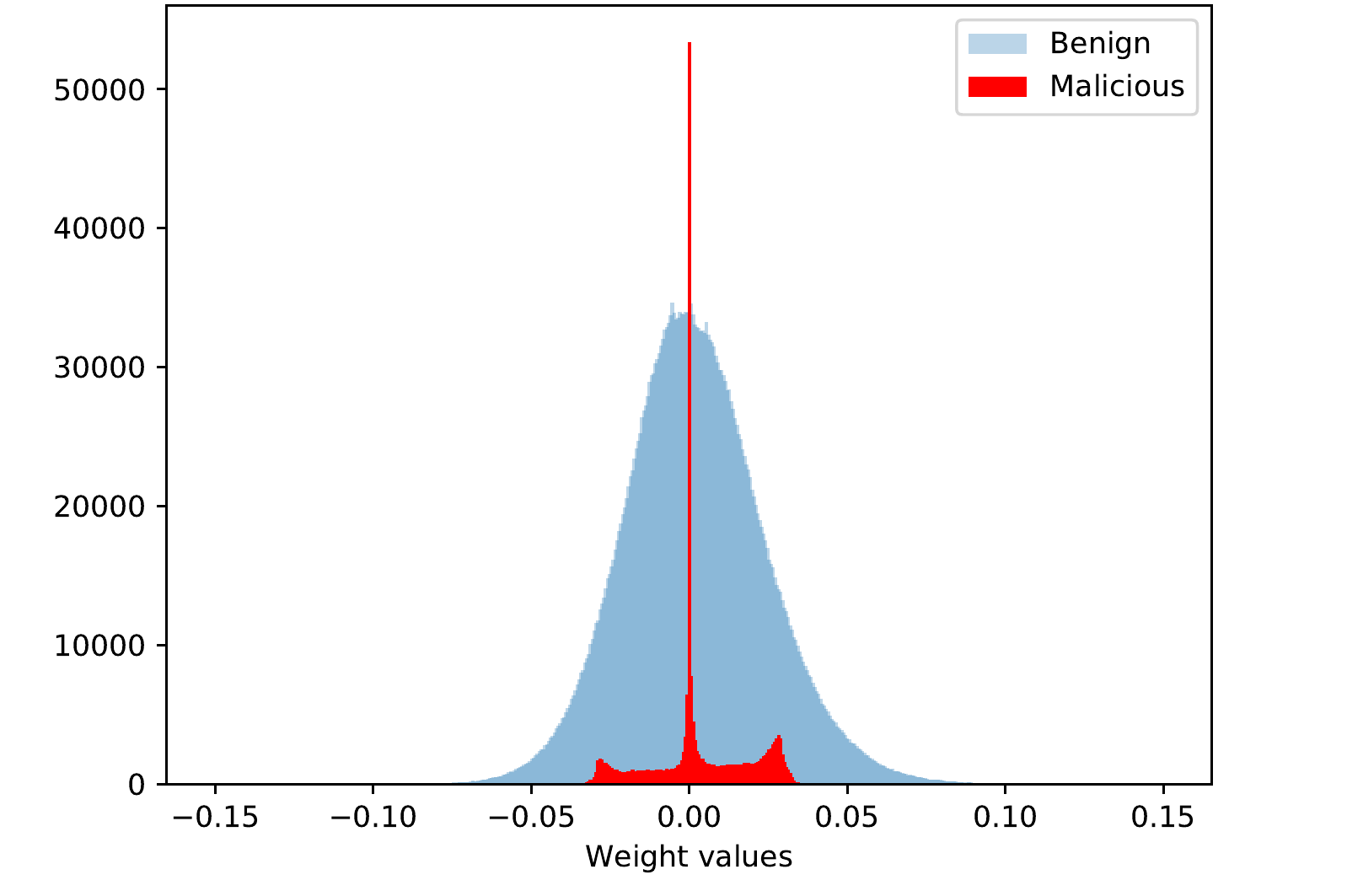}\label{subfig: converge_weights}}
	\caption{\textbf{Targeted model poisoning attack for CNN on Fashion MNIST data.} The total number of agents is $K=10$, including the malicious agents. All agents train their local models for 5 epochs with the appropriate objective.}
	\vspace{-10pt}
\end{figure}

\subsection{Targeted model poisoning for standard federated learning}\label{subsec: baseline}
The adversary can directly optimize the adversarial objective $L(\{\bfx_i,\tau_i\}_{i=1}^r, \hat{\bfw}^t_G)$ with $\hat{\bfw}_G^{t} = \bfw_G^{t-1} +  \alpha_m \bdelta^{t}_m$. However, this setup implies that the optimizer has to account for the scaling factor $\alpha_m$ \emph{implicitly}. In practice, we find that when using a gradient-based optimizer such as SGD, \emph{explicit boosting} is much more effective. The rest of the section focuses on explicit boosting and an analysis of implicit boosting is deferred to Section \ref{appsec: imp_boost} of the Appendix.

\noindent \textbf{Explicit Boosting:} Mimicking a benign agent, the malicious agent can run $E_m$ steps of a gradient-based optimizer starting from $\bfw_G^{t-1}$ to obtain $\tilde{\bfw}_m^t$ which minimizes the loss over $\{\bfx_i,\tau_i\}_{i=1}^r$. The malicious agent then obtains an initial update $\tilde{\bdelta}^{t}_m = \tilde{\bfw}_m^t - \bfw_G^{t-1}$. However, since the malicious agent's update tries to ensure that the model learns labels different from the true labels for the data of its choice ($\mcD_{\text{aux}}$), it has to overcome the effect of scaling, which would otherwise mostly nullify the desired classification outcomes. This happens because the learning objective for all the other agents is very different from that of the malicious agent, especially in the i.i.d. case. The final weight update sent back by the malicious agent is then $\bdelta^{t}_m = \lambda \tilde{\bdelta}^{t}_m$, where $\lambda$ is the factor by which the malicious agent \emph{boosts} the initial update. Note that with $\hat{\bfw}_G^{t} = \bfw_G^{t-1} +  \alpha_m \bdelta^{t}_m$ and $\lambda = \frac{1}{\alpha_m}$, then $\hat{\bfw}_G^{t} = \bfw_m^t$, implying that if the estimation was exact, the global weight vector should now satisfy the malicious agent's objective.

\noindent \textbf{Results:} In the attack with \emph{explicit boosting}, the malicious agent runs $E_m=5$ steps of the Adam optimizer \cite{kingma2014adam} to obtain $\tilde{\bdelta}^{t}_m$, and then boosts it by $\frac{1}{\alpha_m}=K$. The results for the case with $K=10$ for the Fashion MNIST data are shown in Figure \ref{subfig: baseline_attack}. The attack is clearly successful at causing the global model to classify the chosen example in the target class. In fact, after $t=3$, the global model is highly confident in its (incorrect) prediction. Further, the global model converges with good performance on the validation set in spite of the targeted poisoning for 1 example. Results for the Adult Census dataset (Section \ref{appsubsec: census_data}) demonstrate targeted model poisoning is possible across datasets and models. Thus, the explicit boosting attack is able to achieve targeted poisoning in the federated learning setting.

\begin{figure}[t]
	\centering
	\subfloat[Confidence on malicious objective and accuracy on validation data for $\bfw_G^t$. Stealth with respect to accuracy checking is also shown for both the stealthy and targeted model poisoning attacks.]{\resizebox{0.48\textwidth}{!}{\input{latex_plots_new/fmnist_m0_mal_concat_train_dist_oth_.tex}}\label{subfig: concat_train_attack}}
	\hspace{2mm}
	\subfloat[Comparison of weight update distributions for benign and malicious agents]{\includegraphics[width=0.37\textwidth]{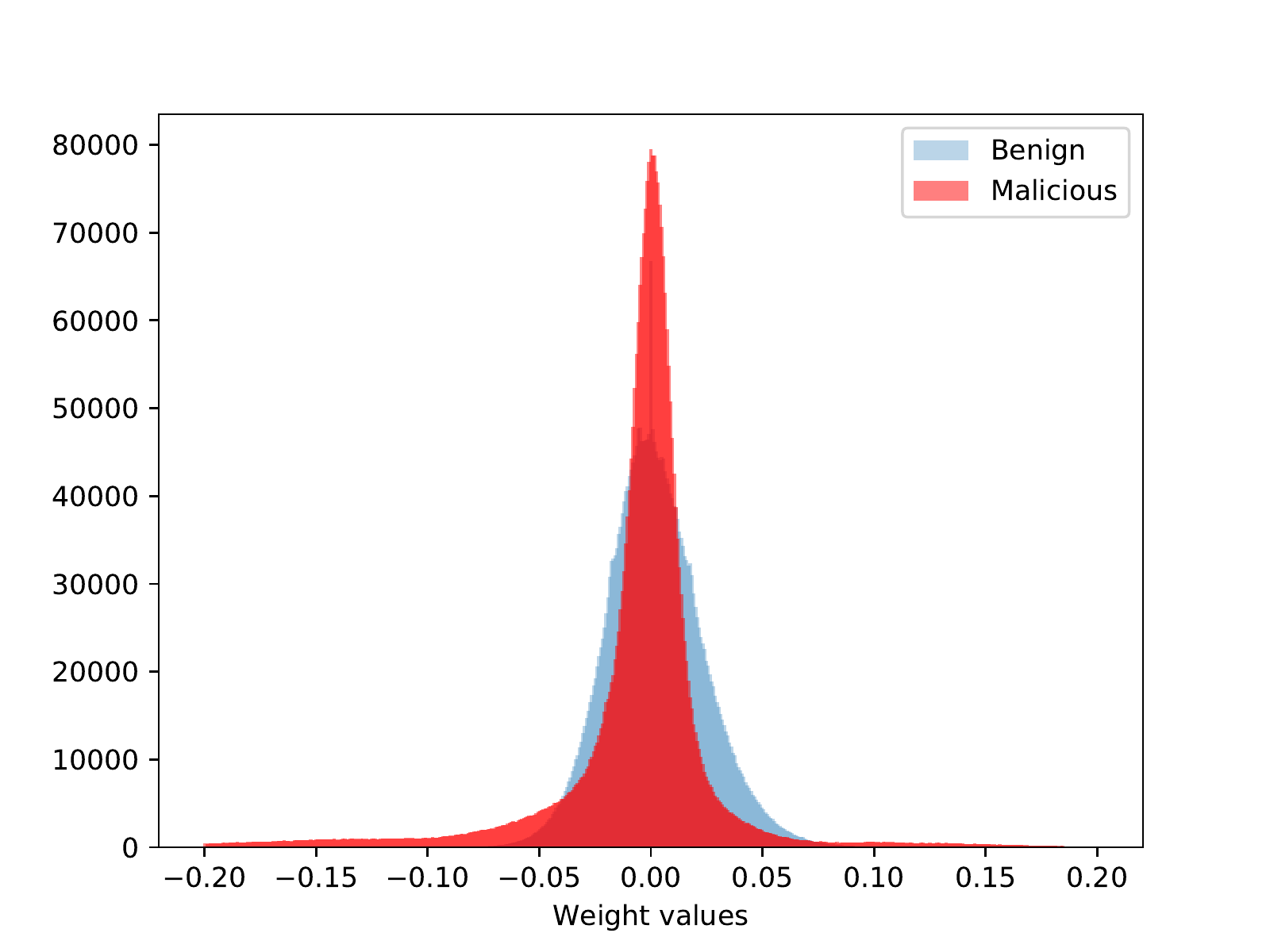}\label{subfig: concat_train_weights}}
	\caption{\textbf{Stealthy model poisoning for CNN on Fashion MNIST}. We use $\lambda=10$ and $\rho=1e^{-4}$ for the malicious agent's objective.}
	\vspace{-13pt}
\end{figure}

 
\noindent \textbf{Performance on stealth metrics:} While the targeted model poisoning attack using explicit boosting does not take stealth metrics into account, it is instructive to study properties of the model update it generates. Compared to the weight update from a benign agent, the update from the malicious agent is much sparser and has a smaller range (Figure \ref{subfig: converge_weights}). In Figure \ref{fig: l2_dist}, the spread of $L_2$ distances between all benign updates and between the malicious update and the benign updates is plotted. For targeted model poisoning, both the minimum and maximum distance away from any of the benign updates keeps decreasing over time steps, while it remains relatively constant for the other agents. In Figure \ref{subfig: concat_train_attack} the accuracy of the malicious model on the validation data (\emph{Val. Acc. Mal (targeted poison)}) is shown, which is much lower than the global model's accuracy. Thus, both accuracy checking and weight update statistics based detection is possible for the targeted model poisoning attack.

\subsection{Stealthy model poisoning} \label{subsec: stealth_train}

\anew{As discussed in Section \ref{subsec: stealth}, there are two properties which the server can use to detect anomalous updates: accuracy on validation data and weight update statistics. In order to maintain stealth with respect to both of these properties, the adversary can add loss terms corresponding to both of those metrics to the model poisoning objective function from Eq. \ref{eq: adv_opt} and improve targeted model poisoning. First, in order to improve the accuracy on validation data, the adversary adds the training loss over the malicious agent's local data shard $\mcD_m$ ($L(\mcD_m, \bfw^t_G)$) to the objective. Since the training data is i.i.d. with the validation data, this will ensure that the malicious agent's update is similar to that of a benign agent in terms of validation loss and will make it challenging for the server to flag the malicious update as anomalous.}

\anew{Second, the adversary needs to ensure that its update is as close as possible to the benign agents' updates in the appropriate distance metric. For our experiments, we use the $\ell_p$ norm with $p=2$. Since the adversary does not have access to the updates for the current time step $t$ that are generated by the other agents, it constrains $\bm{\delta}^{t}_m$ with respect to $\bar{\bm{\delta}}^{t-1}_{\text{ben}}=\sum_{i \in [k]\setminus m} \alpha_i \bm{\delta}^{t-1}_i$, which is the average update from all the other agents for the previous iteration, which the malicious agent has access to. Thus, the adversary adds $\rho\|\bm{\delta}^{t}_m-\bar{\bm{\delta}}^{t-1}_{\text{ben}}\|_2$ to its objective as well. We note that the addition of the training loss term is not sufficient to ensure that the malicious weight update is close to that of the benign agents since there could be multiple local minima with similar loss values. Overall, the adversarial objective then becomes:
\begin{equation}
\begin{split}
\argmin_{\bdelta^{t}_m} \lambda L(\{\bfx_i,\tau_i\}_{i=1}^r, \hat{\bfw}^t_G) &+ L(\mcD_m, \bfw^t_m)+ \rho\|\bm{\delta}^{t}_m-\bar{\bm{\delta}}^{t-1}_{\text{ben}}\|_2
\end{split}
\end{equation}

Note that for the training loss, the optimization is just performed with respect to $\bfw_m^t=\bfw_G^{t-1}+\bdelta^{t}_m$, as a benign agent would do. Using explicit boosting, $\hat{\bfw}^t_G$ is replaced by $\bfw_m^t$ as well so that only the portion of the loss corresponding to the malicious objective gets boosted by a factor $\lambda$. }

\begin{figure}[t]
	\centering
	\subfloat[Confidence on malicious objective and accuracy on validation data for $\bfw_G^t$. Stealth with respect to accuracy checking is also shown.]{\resizebox{0.48\textwidth}{!}{\input{latex_plots_new/fmnist_m0_mal_alternate_train_dist_oth_.tex}}}
	\hspace{2mm}
	\subfloat[Comparison of weight update distributions for benign and malicious agents]{\includegraphics[width=0.37\textwidth]{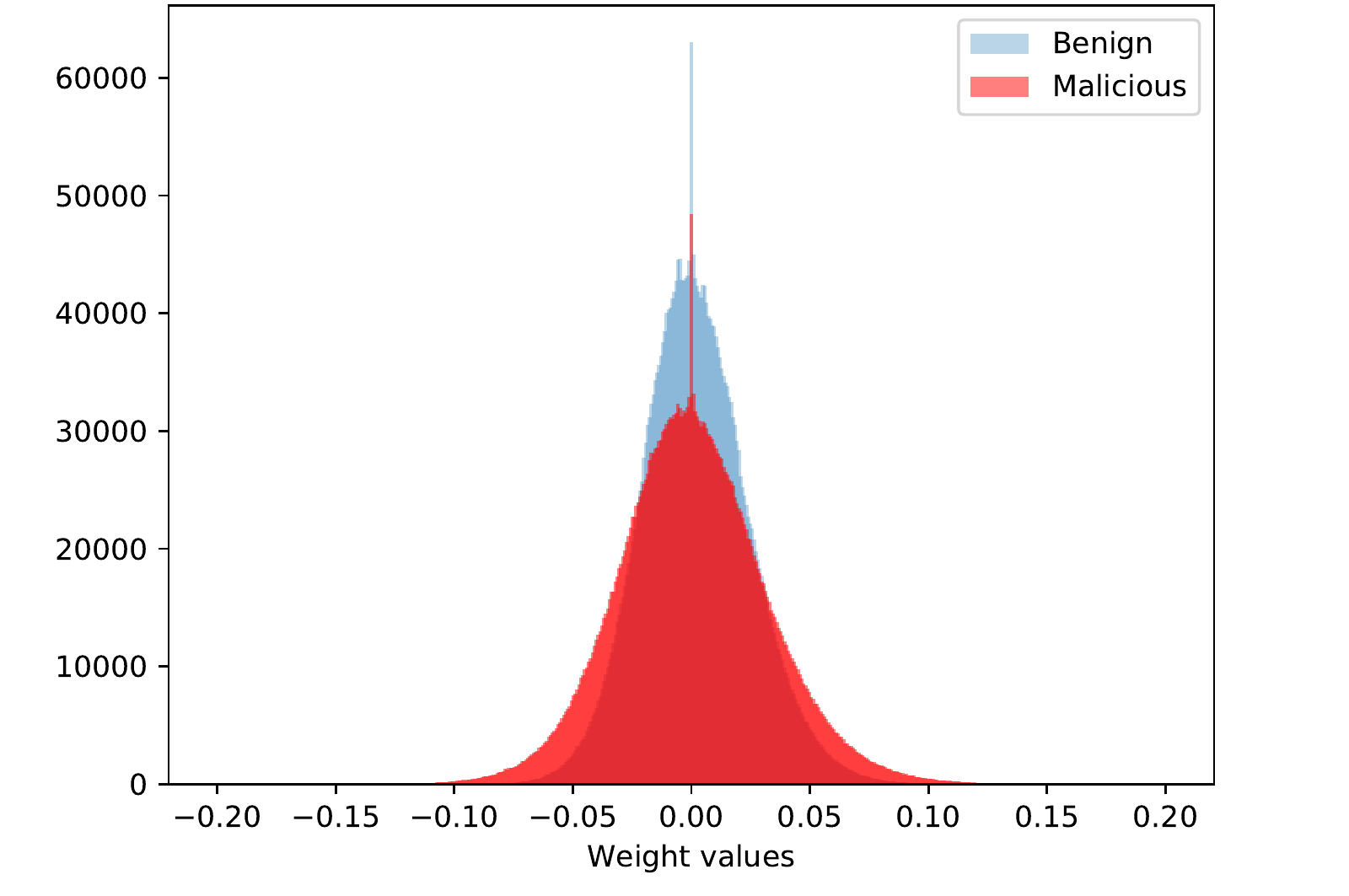}\label{subfig: alt_min_weights}}
	\caption{\textbf{Alternating minimization attack with distance constraints for CNN on Fashion MNIST data}. We use $\lambda=10$ and $\rho=1e^{-4}$. The number of epochs used by the malicious agent is $E_m=10$ and it runs $10$ steps of the stealth objective for every step of the malicious objective.}
	\label{subfig: alt_min_attack}
	\vspace{-10pt}
\end{figure}

\begin{wrapfigure}{r}{0.5\textwidth}
	\resizebox{0.47\textwidth}{!}{\input{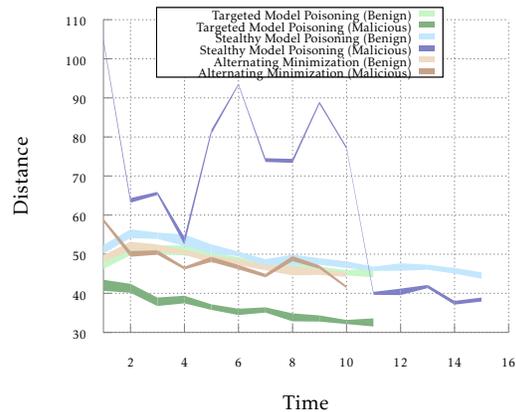}}
	\vspace{-5pt}
	\caption{\textbf{Range of $\ell_2$ distances between all benign agents and between the malicious agent and the benign agents.}}
	\label{fig: l2_dist}
	\vspace{-15pt}
\end{wrapfigure}

\noindent \textbf{Results and effect on stealth:} From Figure \ref{subfig: concat_train_attack}, it is clear that the stealthy model poisoning attack is able to cause targeted poisoning of the global model. We set the accuracy threshold $\gamma_t$ to be 10\% which implies that the malicious model is chosen for 10 iterations out of 15. This is in contrast to the targeted model poisoning attack which never has validation accuracy within 10\% of the global model. Further, the weight update distribution for the stealthy poisoning attack (Figure \ref{subfig: concat_train_weights}) is similar to that of a benign agent, owing to the additional terms in the loss function. Finally, in Figure \ref{fig: l2_dist}, we see that the range of $\ell_2$ distances for the malicious agent $R_m$ is close, according to Eq. \ref{eqn: range_constraint}, to that between benign agents.

Concurrent work on model poisoning boosts the entire update (instead of just the malicious loss component as we do) when the global model is close to convergence in an attempt to perform model replacement \cite{bagdasaryan2018backdoor} but this strategy is ineffective when the model has not converged.

\subsection{Alternating minimization for improved model poisoning} \label{subsec: alt_min}
While the stealthy model poisoning attack ensures targeted poisoning of the global model while maintaining stealth according to the two conditions required, it does not ensure that the malicious agent's update is chosen in every iteration. To achieve this, we propose an \emph{alternating minimization attack strategy} which decouples the targeted objective from the stealth objectives, providing finer control over the relative effect of the two objectives. It works as follows for iteration $t$. For each epoch $i$, the adversarial objective is first minimized starting from $\bfw^{i-1,t}_m$, giving an update vector $\tilde{\bdelta}^{i,t}_m$. This is then boosted by a factor $\lambda$ and added to $\bfw^{i-1,t}_m$. Finally, the stealth objective for that epoch is minimized starting from $\tilde{\bfw}^{i,t}_m = \bfw^{i-1,t}_m+\lambda \tilde{\bdelta}^{i,t}_m$, providing the malicious weight vector $\bfw^{i,t}_m$ for the next epoch. The malicious agent can run this alternating minimization until both the adversarial and stealth objectives have sufficiently low values. Further, the independent minimization allows for each objective to be optimized for a different number of steps, depending on which is more difficult in achieve. In particular, we find that optimizing the stealth objective for a larger number of steps each epoch compared to the malicious objective leads to better stealth performance while maintaining targeted poisoning.

\noindent \textbf{Results and effect on stealth:} The adversarial objective is achieved at the global model with high confidence starting from time step $t=2$ and the global model converges to a point with good performance on the validation set. This attack can bypass the accuracy checking method as the accuracy on validation data of the malicious model is close to that of the global model.In Figure \ref{fig: l2_dist}, we can see that the distance spread for this attack closely follows and even overlaps that of benign updates throughout, thus achieving complete stealth with respect to both properties.

\section{Attacking Byzantine-resilient aggregation}
\label{sec: byzantine_attack}
\begin{wrapfigure}{r}{0.5\textwidth}
	\centering
	\resizebox{0.47\textwidth}{!}{\input{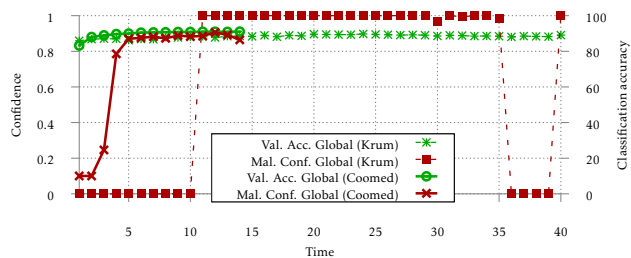}}
	\caption{\textbf{Model poisoning attacks with Byzantine resilient aggregation mechanisms.} We use targeted model poisoning for \textsf{coomed} and alternating minimization for Krum.}
	\label{fig: byzantine_plots}
	\vspace{-10pt}
\end{wrapfigure}

There has been considerable recent work that has proposed gradient aggregation mechanisms for distributed learning that ensure convergence of the global model \cite{peva17,yudong17,mhamdi2018hidden,chen18,yin2018byzantine}. However, the aim of the Byzantine adversaries considered in this line of work is to ensure convergence to ineffective models, i.e. models with poor classification performance. The goal of the adversary we consider is targeted model poisoning, which implies convergence to an effective model on the test data. This difference in objectives leads to the lack of robustness of these Byzantine-resilient aggregation mechanisms against our attacks. We consider the aggregation mechanisms Krum \cite{peva17} and coordinate-wise median \cite{yin2018byzantine} for our evaluation, both of which are provably Byzantine-resilient and converge under appropriate conditions on the loss function. Both aggregation mechanisms are also efficient. Note that in general, these conditions do not hold for neural networks so the guarantees are only empirical.

\subsection{Krum} 
Given $n$ agents of which $f$ are Byzantine, Krum requires that $n\geq 2f+3$. At any time step $t$, updates $\left( \bm{\delta}_1^t, \ldots, \bm{\delta}_n^t  \right)$ are received at the server. For each $\bm{\delta}_i^t$, the $n-f-2$ closest (in terms of $L_p$ norm) other updates are chosen to form a set $C_i$ and their distances added up to give a score $S(\bm{\delta}_i^t) = \sum_{\bm{\delta} \in C_i} \|\bm{\delta}_i^t-\bm{\delta}\|$. Krum then chooses $\bm{\delta}_{\textbf{krum}}=\bm{\delta}_i^t$ with the lowest score to add to $\bfw_i^t$ to give $\bfw_i^{t+1}=\bfw_i^t + \bm{\delta}_{\textbf{krum}}$. In Figure \ref{fig: byzantine_plots}, we see the effect of the alternating minimization attack on Krum with a boosting factor of $\lambda=2$ for a federated learning setup with 10 agents. Since there is no need to overcome the constant scaling factor $\alpha_m$, the attack can use a much smaller boosting factor $\lambda$ than the number of agents to ensure model poisoning. The malicious agent's update is chosen by Krum for 26 of 40 time steps which leads to the malicious objective being met. Further, the global model converges to a point with good performance as the malicious agent has added the training loss to its stealth objective. We note that with the use of targeted model poisoning, we can cause Krum to converge to a model with poor performance as well (see Appendix \ref{appsubsec: krum}).

\subsection{Coordinate-wise median} 
Given the set of updates $\{\bdelta^t_i\}_{i=1}^k$ at time step $t$, the aggregate update is $\bar{\bdelta}^t \coloneqq \textsf{coomed}\{\{\bdelta^t_i\}_{i=1}^k\}$, which is a vector with its $j^{\text{th}}$ coordinate $\bar{\bdelta}^t(j) = \textsf{med}\{\bdelta^t_i(j)\}$, where $\textsf{med}$ is the 1-dimensional median. Using targeted model poisoning with a boosting factor of $\lambda=1$, i.e. no boosting, the malicious objective is met with confidence close to 0.9 for 11 of 14 time steps (Figure \ref{fig: byzantine_plots}). We note that in this case, unlike with Krum, there is convergence to an effective global model. We believe this occurs due to the fact that coordinate-wise median does not simply pick one of the updates to apply to the global model and does indeed use information from all the agents while computing the new update. Thus, \emph{model poisoning attacks are effective against two completely different Byzantine-resilient aggregation mechanisms}.

\section{Improving attack performance through estimation}
\label{sec: poison_estimation}
In this section, we look at how the malicious agent can choose a better estimate for the effect of the other agents' updates at each time step that it is chosen. In the case when the malicious agent is not chosen at every time step, this estimation is made challenging by the fact that it may not have been chosen for many iterations. 

\subsection{Estimation setup}  \label{subsec: est_setup}
The malicious agent's goal is to choose an appropriate estimate for $\bm{\delta}^t_{[k] \setminus m} = \sum_{i \in [k] \setminus m} \alpha_i \bdelta^{t}_i$, i.e. for the effects of the other agents at time step $t$. When the malicious agent is chosen at time $t$, the following information is available to them from the previous time steps they were chosen: i) Global parameter vectors $\bfw_G^{t_0} \ldots, \bfw_G^{t-1}$; ii) Malicious weight updates $\bm{\delta}_m^{t_0} \ldots, \bm{\delta}_m^t$; and iii) Local training data shard $\mcD_m$, where $t_0$ is the first time step at which the malicious agent is chosen. Given this information, the malicious agent computes an estimate $\hat{\bm{\delta}}^t_{[k] \setminus m}$ which it can use to correct for the effect of other agents in two ways:\\

\noindent \textbf{1. Post-optimization correction:} In this method, once the malicious agent computes its weight update $\bm{\delta}_m^t$, it subtracts $\lambda \hat{\bm{\delta}}^t_{[k] \setminus m}$ from it before sending it to the server. If $\hat{\bm{\delta}}^t_{[k] \setminus m}=\bm{\delta}^t_{[k] \setminus m}$ and $\lambda=\frac{1}{\alpha_m}$, this will negate the effects of other agents.\\

\noindent \noindent \textbf{2. Pre-optimization correction:} Here, the malicious agent assumes that $\hat{\bfw}_G^{t} = \bfw_G^{t-1} + \hat{\bm{\delta}}^t_{[k] \setminus m} + \alpha_m \bdelta^{T+1}_m$. In other words, the malicious agent optimizes for $\bdelta_m^{t}$ assuming it has an accurate estimate of the other agents' updates. For attacks which use explicit boosting, this involves starting from $\bfw_G^{t-1} + \hat{\bm{\delta}}^t_{[k] \setminus m}$ instead of just $\bfw_G^{t-1} $. 

\subsection{Estimation strategies and results}  \label{subsec: est_strat}
When the malicious agent is chosen at time step $t$ \footnote{If they are chosen at $t=0$ or $t$ is the first time they are chosen, there is no information available regarding the other agents' updates}, information regarding the probable updates from the other agents can be obtained from the previous time steps at which the malicious agent was chosen. 

\subsubsection{Previous step estimate} 
In this method, the malicious agent's estimate $\hat{\bm{\delta}}^t_{[k] \setminus m}$ assumes that the other agents' cumulative updates were the same at each step since $t'$ (the last time step at which at the malicious agent was chosen), i.e.
$
\hat{\bm{\delta}}^t_{[k] \setminus m} = \frac{\bfw_G^t-\bfw_G^{t'}-\bm{\delta}^{t'}_m}{t-t'}.
$
In the case when the malicious agent is chosen at every time step, this reduces to $\hat{\bm{\delta}}^t_{[k] \setminus m} =\bm{\delta}^{t-1}_{[k] \setminus m}$. This estimate can be applied to both the pre- and post-optimization correction methods.

\begin{table}
	\centering
	\begin{tabular}{c|cc|cc}
		Attack & \multicolumn{2}{c|}{ \begin{tabular}{c} Targeted \\ Model Poisoning \end{tabular}} & \multicolumn{2}{c}{\begin{tabular}{c} Alternating \\ Minimization \end{tabular}} \\ \hline
		Estimation & None & Previous step & None & Previous step \\ \hline
		$t=2$ & 0.63 & 0.82 & 0.17 & 0.47 \\ 
		$t=3$ & 0.93 & 0.98 & 0.34 & 0.89 \\ 
		$t=4$ & 0.99 & 1.0  &  0.88 & 1.0 \\ \bottomrule
	\end{tabular}
	\caption{Comparison of confidence of targeted misclassification with and without the use of previous step estimation for the targeted model poisoning and alternating minimization attacks.}
	\label{tab: estimation}
	\vspace{-10pt}
\end{table}

\subsubsection{Results} 
Attacks using previous step estimation with the pre-optimization correction are more effective at achieving the adversarial objective for both the targeted model poisoning and alternating minimization attacks. In Table \ref{tab: estimation}, we can see that the global model misclassifies the desired sample with a higher confidence when using previous step estimation in the first few iterations. We found that using post-optimization correction was not effective, leading to both lower attack success and affecting global model convergence.


\section{Discussion}
\label{sec: discussion}
\subsection{Model poisoning vs. data poisoning}
\label{subsec: data_poison}
In this section, we elucidate the differences between model poisoning and data poisoning both qualitatively and quantitatively. Data poisoning attacks largely fall in two categories: clean-label \cite{munoz2017towards,koh2017understanding} and dirty-label \cite{chen17,gu2017badnets,liu2017trojaning}. Clean-label attacks assume that the adversary \emph{cannot} change the label of any training data as there is a process by which data is certified as belonging to the correct class and the poisoning of data samples has to be imperceptible. On the other hand, to carry out dirty-label poisoning, the adversary just has to introduce a number of copies of the data sample it wishes to mis-classify with the desired target label into the training set since there is no requirement that a data sample belong to the correct class. Dirty-label data poisoning has been shown to achieve high-confidence targeted misclassification for deep neural networks with the addition of around 50 poisoned samples to the training data  \cite{chen17}. 

\subsubsection{Dirty-label data poisoning in federated learning} 
In our comparison with data poisoning, we use the dirty-label data poisoning framework for two reasons. First, federated learning operates under the assumption that data is never shared, only learned models. Thus, the adversary is not concerned with notions of imperceptibility for data certification. Second, clean-label data poisoning assumes access at train time to the global parameter vector, which is absent in the federated learning setting. Using the same experimental setup as before (CNN on Fashion MNIST data, $10$ agents chosen every time step), we add copies of the sample that is to be misclassified to the training set of the malicious agent with the appropriate target label. We experiment with two settings. In the first, we add multiple copies of the same sample to the training set. In the second, we add a small amount of random uniform noise to each pixel \cite{chen17} when generating copies. We observe that even when we add $1000$ copies of the sample to the training set, the \emph{data poisoning attack is completely ineffective at causing targeted poisoning in the global model}. This occurs due to the fact that malicious agent's update is scaled, which again underlies the importance of boosting while performing model poisoning. We note also that if the update generated using data poisoning is boosted, it affects the performance of the global model as the entire update is boosted, not just the malicious part. Thus, model poisoning attacks are much more effective than data poisoning in the federated learning setting. 


\subsection{Interpreting poisoned models}
\label{subsec: poison_interpret}
\begin{figure}
	\subfloat[Decision visualization for auxiliary data sample on a model with trained with $10$ benign agents.]
	{\includegraphics[width=0.95\textwidth]{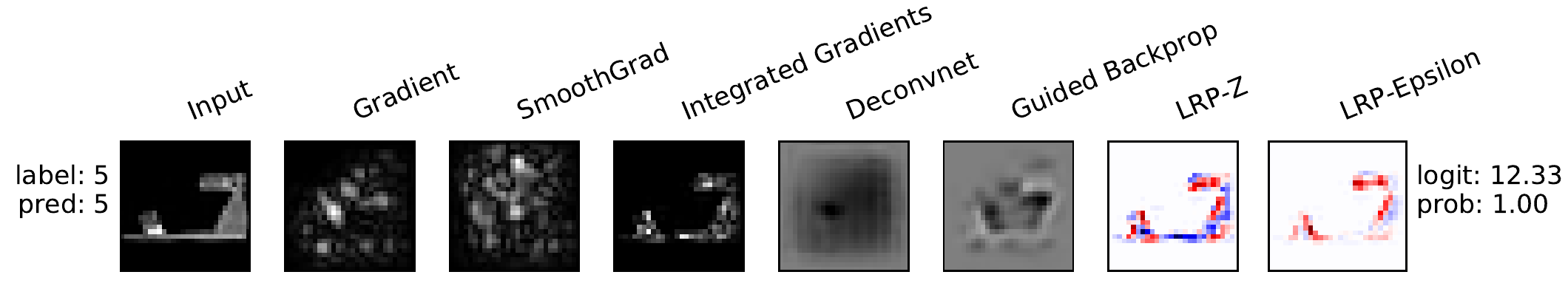}}
	\label{subfig: ben_on_mal_data}
	\subfloat[Decision visualization for auxiliary data sample on a model trained with $9$ benign agents and a single malicious agent.]
	{\includegraphics[width=0.95\textwidth]{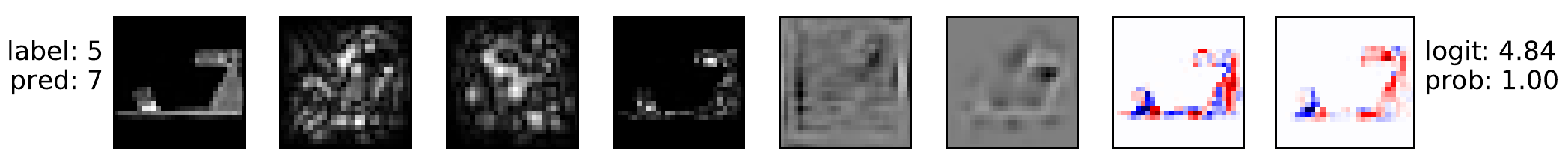}}
	\label{subfig: mal_on_mal_data}
	\caption{\textbf{Decision visualizations for benign and malicious models for a CNN on the Fashion MNIST data.}}
	\label{fig: interpret_mal}
\end{figure}

Neural networks are often treated as black boxes with little transparency into their internal representation or understanding of the underlying basis for their decisions. Interpretability techniques are designed to alleviate these problems by analyzing various aspects of the network. These include (i) identifying the relevant features in the input pixel space for a particular decision via Layerwise Relevance Propagation (LRP) techniques (\cite{montavon15}); (ii) visualizing the association between neuron activations and image features (Guided Backprop (\cite{springenberg14}), DeConvNet (\cite{zeiler14})); (iii) using gradients for attributing prediction scores to input features (e.g., Integrated Gradients (\cite{sundararajan17}), or generating sensitivity and saliency maps (SmoothGrad (\cite{smilkov17}), Gradient Saliency Maps (\cite{simonyan13})) and so on. The semantic relevance of the generated visualization, relative to the input, is then used to explain the model decision.

These interpretability techniques, in many ways, provide insights into the internal feature representations and working of a neural network. Therefore, we used a suite of these techniques to try and discriminate between the behavior of a benign global model and one that has been trained to satisfy the adversarial objective of misclassifying a single example. Figure~\ref{fig: interpret_mal} compares the output of the various techniques for both the benign and malicious models on a random auxiliary data sample. Targeted perturbation of the model parameters coupled with tightly bounded noise ensures that the internal representations, and relevant input features used by the two models, for the same input, are almost visually imperceptible. This further exposes the fragility of interpretability methods \cite{adebayo2018sanity}.



\bibliographystyle{abbrv}
\bibliography{fedlearnpoison.bib}

\appendix

\begin{wrapfigure}{r}{0.5\textwidth}
	\vspace{-20pt}
	\hspace{5pt}
	\resizebox{0.45\textwidth}{!}{\input{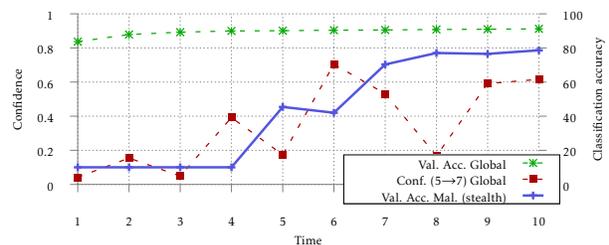}}
	\caption{\textbf{Targeted model poisoning with implicit boosting.} The number of agents was $k=10$ with a CNN on Fashion MNIST data.}
	\label{fig: imp_boost}
	\vspace{-10pt}
\end{wrapfigure}

\section{Implicit Boosting}\label{appsec: imp_boost}
While the loss is a function of a weight vector $\bfw$, we can use the chain rule to obtain the gradient of the loss with respect to the weight update $\bm{\delta}$, i.e. $\nabla_{\bm{\delta}}L = \alpha_m \nabla_{\bfw}L$. Then, initializing $\bm{\delta}$ to some appropriate $\bm{\delta}_{\text{ini}}$, the malicious agent can directly minimize with respect to $\bm{\delta}$. However, the baseline attack using \emph{implicit boosting} (Figure \ref{fig: imp_boost}) is much less successful than the explicit boosting baseline, with the adversarial objective only being achieved in 4 of 10 iterations. Further, it is computationally more expensive, taking an average of $2000$ steps to converge at each time step, which is about $4\times$ longer than a benign agent. Since consistently delayed updates from the malicious agent might lead to it being dropped from the system in practice, we focused on explicit boosting attacks throughout.

\section{Further results}

\subsection{Results on Adult Census dataset}\label{appsubsec: census_data}
\begin{figure}
	\centering
	\subfloat[Targeted model poisoning]{\resizebox{0.47\columnwidth}{!}{\input{latex_plots_new/census_baseline_.tex}}\label{subfig: converge_census}}
	\hspace{5mm}
	\subfloat[Comparison of weight update distributions for targeted model poisoning]{\includegraphics[width=0.37\textwidth]{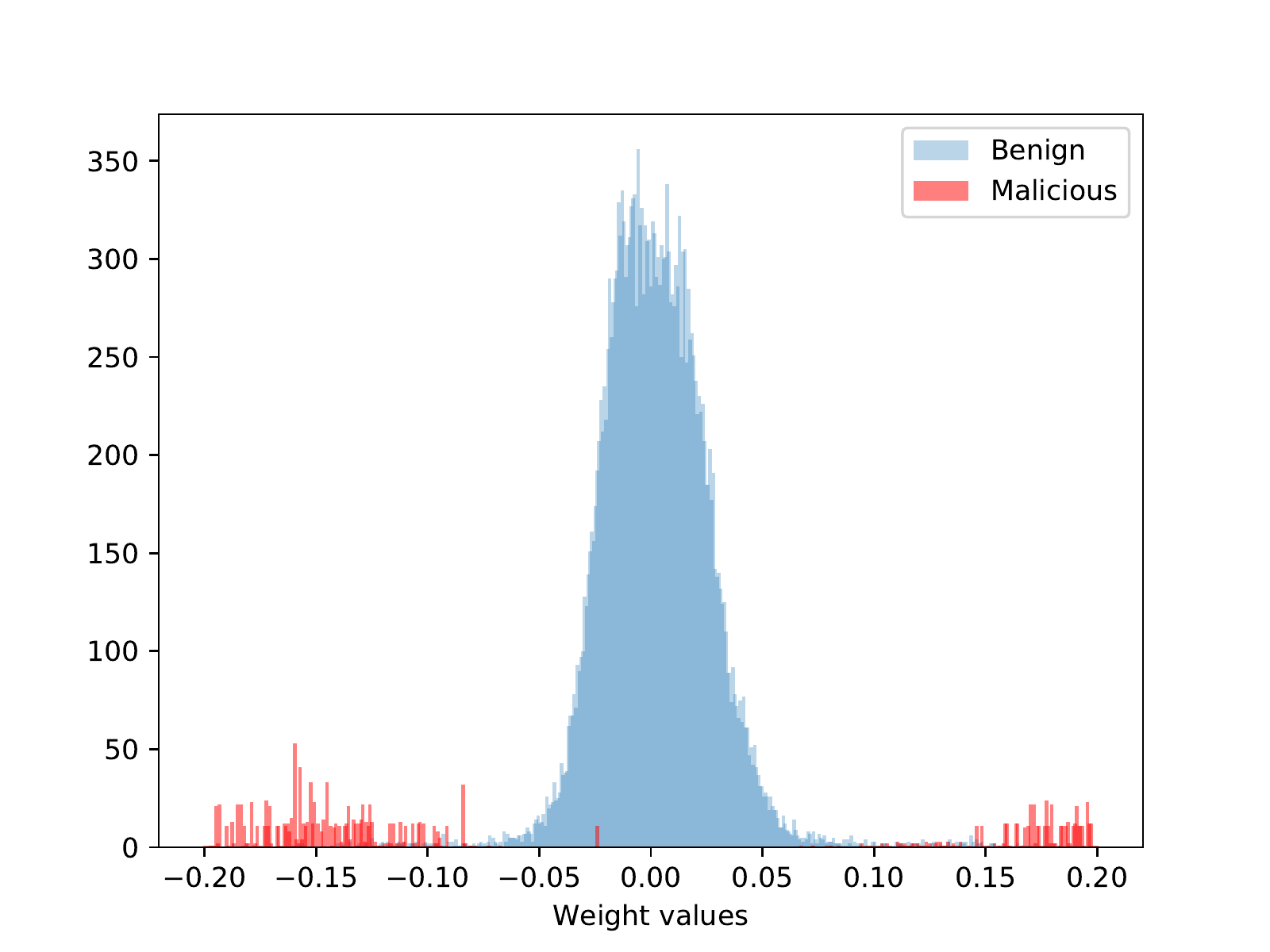}\label{subfig: converge_census_weights}}
	\hspace{0mm}
	\subfloat[Stealthy model poisoning with $\lambda=20$ and $\rho=1e^{-4}$]{\resizebox{0.47\columnwidth}{!}{\input{latex_plots_new/census_stealthy_.tex}}\label{subfig: stealthy_census}}
	\hspace{5mm}
	\subfloat[Comparison of weight update distributions for stealthy model poisoning]{\includegraphics[width=0.37\textwidth]{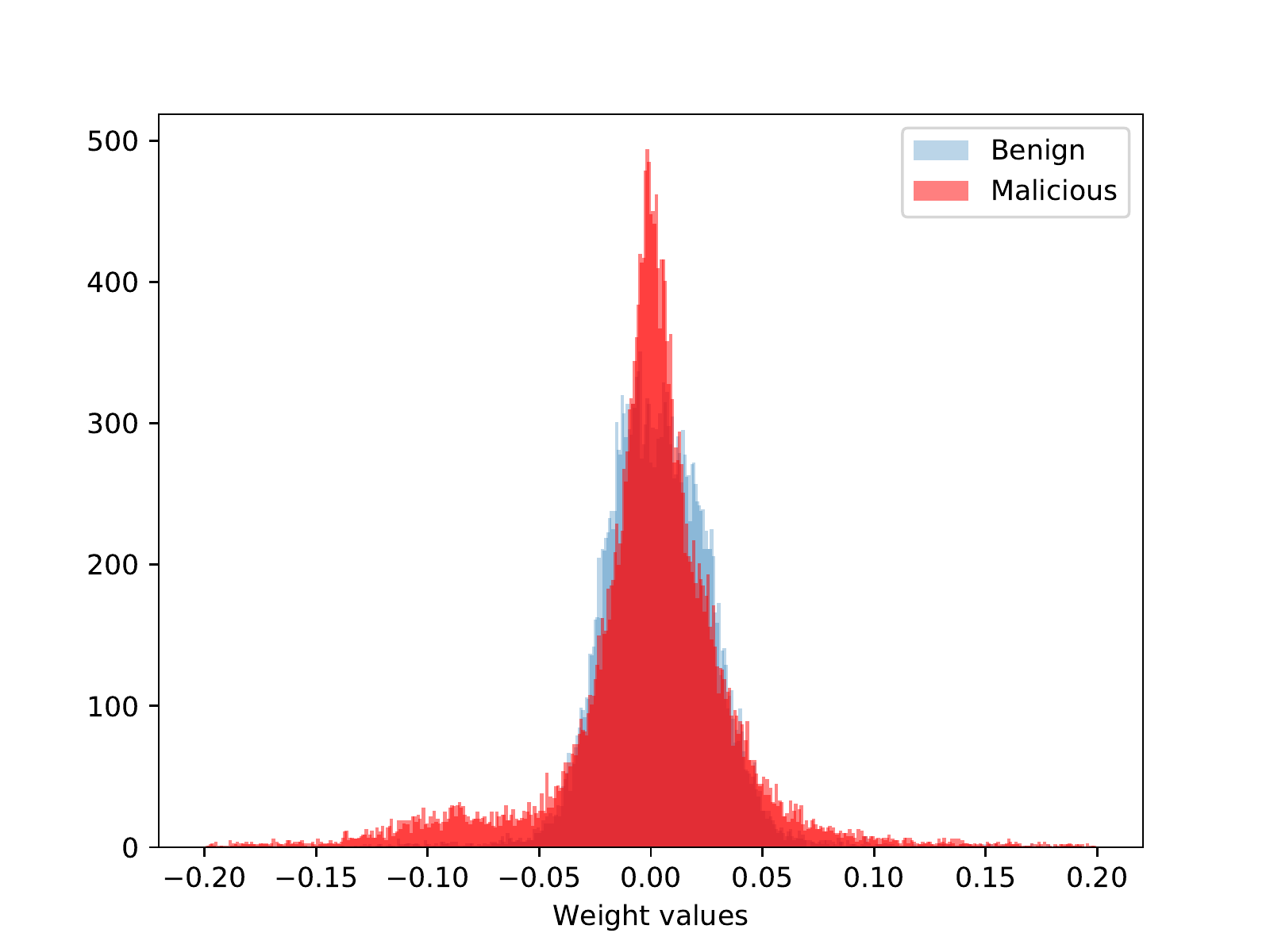}\label{subfig: stealthy_census_weights}}
	\hspace{0mm}
	\subfloat[Alternating minimization with $\lambda=20$ and $\rho=1e^{-4}$ and $10$ epochs for the malicious agent]{\resizebox{0.47\columnwidth}{!}{\input{latex_plots_new/census_alternate_.tex}}\label{subfig: alt_min_census}}
	\hspace{5mm}
	\subfloat[Comparison of weight update distributions for alternating minimization]{\includegraphics[width=0.37\textwidth]{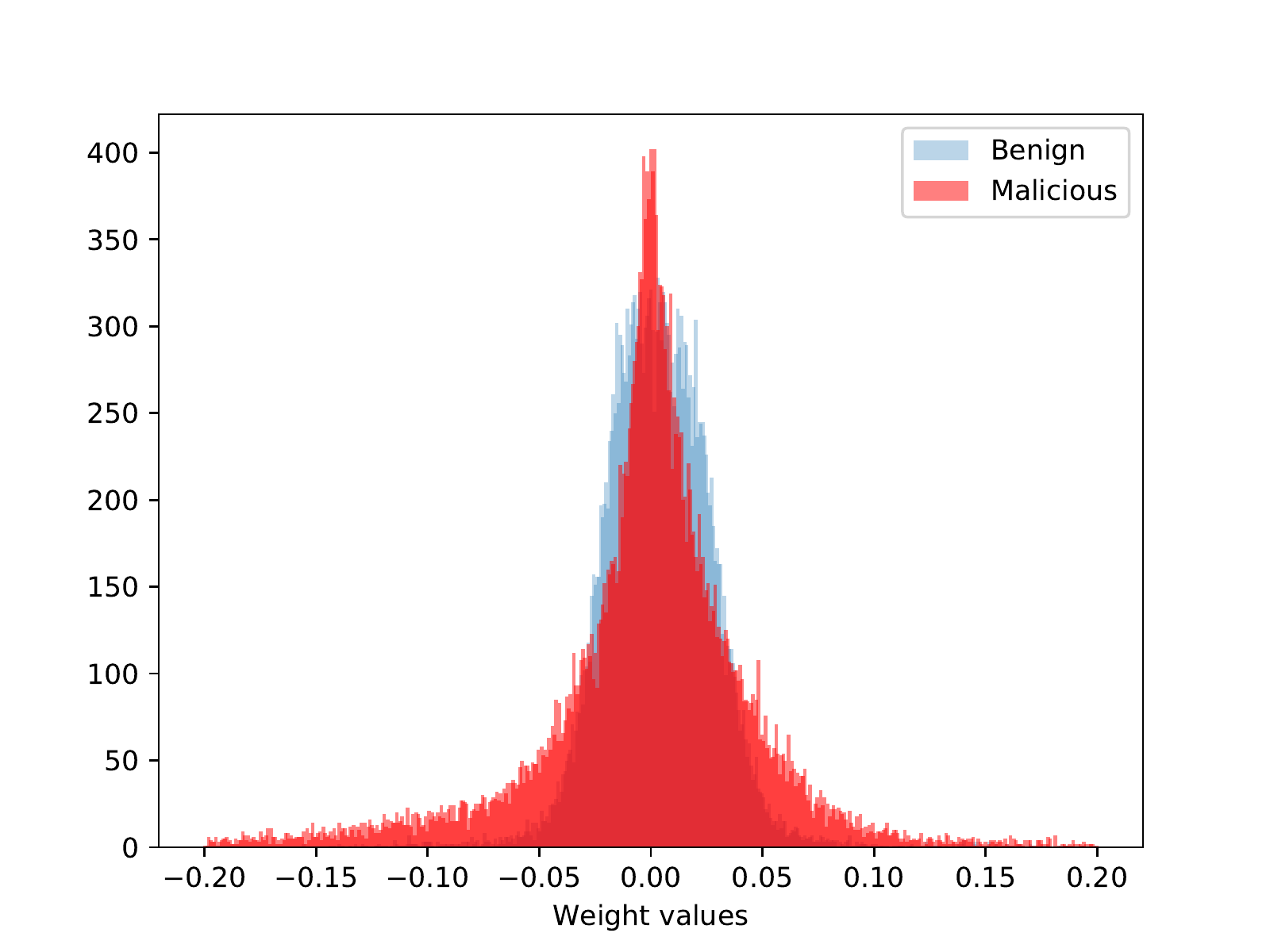}\label{subfig: alt_min_census_weights}}
	\caption{\textbf{Attacks on a fully connected neural network on the Census dataset.}}
	\label{fig: census_plots}
	\vspace{-10pt}
\end{figure}

Results for the 3 different attack strategies on the Adult Census dataset (Figure \ref{fig: census_plots}) confirm the broad conclusions we derived from the Fashion MNIST data. The baseline attack is able to induce high confidence targeted misclassification for a random test example but affects performance on the benign objective, which drops from 84.8\% in the benign case to just around 80\%. The alternating minimization attack is able to ensure misclassification with a confidence of around 0.7 while maintaining 84\% accuracy on the benign objective.

\subsection{Multiple instance poisoning}\label{appsubsec: multiple}
\begin{figure}
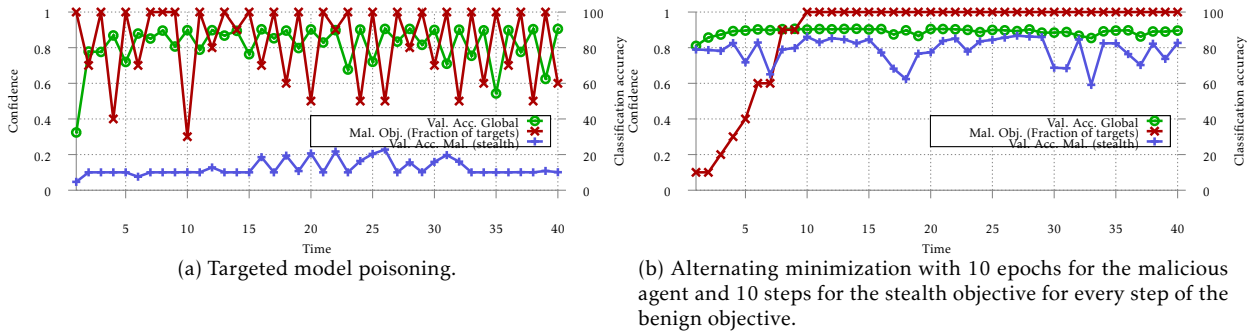

	\centering
	\subfloat[Targeted model poisoning.]{\resizebox{0.47\columnwidth}{!}{\input{latex_plots_new/fmnist_m0_mal_mult10_converge_.tex}}\label{subfig: mult_10_converge}}
	\hspace{3mm}
	\subfloat[Alternating minimization with $10$ epochs for the malicious agent and $10$ steps for the stealth objective for every step of the benign objective.]{	\resizebox{0.47\columnwidth}{!}{\input{latex_plots_new/fmnist_m0_mal_mult10_converge_ls10_ext10_.tex}}\label{subfig: mult_10_alt_min}}
	\caption{\textbf{Attacks with multiple targets ($r=10$) for a CNN on the Fashion MNIST data.}}
	\vspace{-10pt}
\end{figure}

For completeness, we provide results for the case with $r=10$, i.e. the case when the malicious agent wishes to classify 10 different examples in specific target classes. These results are given Figures \ref{subfig: mult_10_converge} (targeted model poisoning) and \ref{subfig: mult_10_alt_min} (Alternating minimization with stealth). While targeted model poisoning is able to induce targeted misclassification, it has an adverse impact on the global model's accuracy. This is countered by the alternating minimization attack, which ensures that the global model converges while still meeting the malicious objective.

\subsection{Randomized agent selection}\label{appsubsec: rand_agent}
\begin{figure}
	\centering
	\subfloat[Targeted model poisoning with $\lambda=100$.]{\resizebox{0.47\columnwidth}{!}{\input{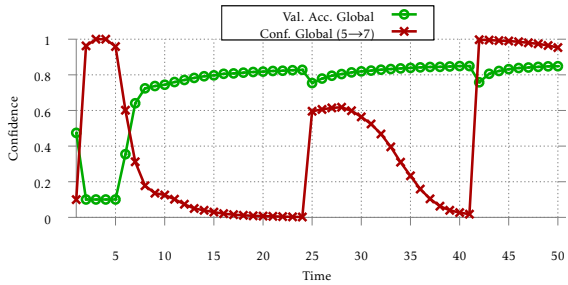}}\label{subfig: single_100_converge}}
	\hspace{3mm}
	\subfloat[Alternating minimization with $\lambda=100$, $100$ epochs for the malicious agent and $10$ steps for the stealth objective for every step of the benign objective.]{	\resizebox{0.47\columnwidth}{!}{\input{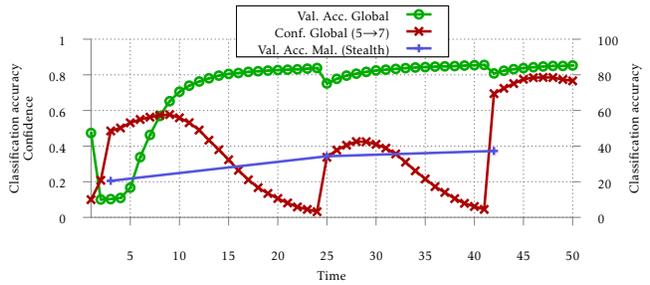}}\label{subfig: single_100_alt_min}}
	\caption{\textbf{Attacks on federated learning in a setting with $K=100$ and a single malicious agent for a CNN on the Fashion MNIST data.}}
	\vspace{-10pt}
\end{figure}

When the number of agents increases to $k=100$, the malicious agent is not selected in every step. Further, the size of $|\mcD_m|$ decreases, which makes the benign training step in the alternating minimization attack more challenging. The challenges posed in this setting are reflected in Figure \ref{subfig: single_100_converge}, where although targeted model poisoning is able to introduce a targeted backdoor, it is not present for every step as there are steps where only benign agents provide updates. Nevertheless, targeted model poisoning is effective overall, with the malicious objective achieved along with convergence of the global model at the end of training. The alternating minimization attack strategy with stealth (Figure \ref{subfig: single_100_alt_min}) is also able to introduce the backdoor, as well as increase the classification accuracy of the malicious model on test data. However, the improvement in performance is limited by the paucity of data for the malicious agent. It is an open question if data augmentation could help improve this accuracy.


\subsection{Bypassing Byzantine-resilient aggregation mechanisms}\label{appsubsec: krum}
\begin{figure}
	\centering
	\subfloat[Targeted model poisoning with $\lambda=2$ against Krum.]{\resizebox{0.47\columnwidth}{!}{\input{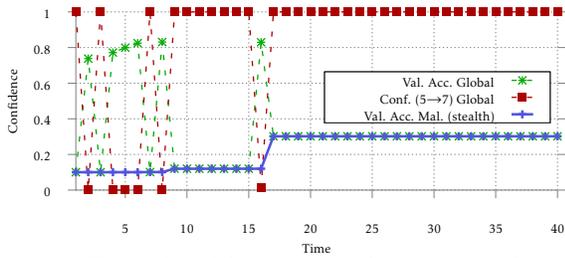}}\label{subfig: baseline_krum}}
	\hspace{4mm}
	\subfloat[Alternating minimization attack with $\lambda=2$ against \textsf{coomed}.]{\resizebox{0.47\columnwidth}{!}{\input{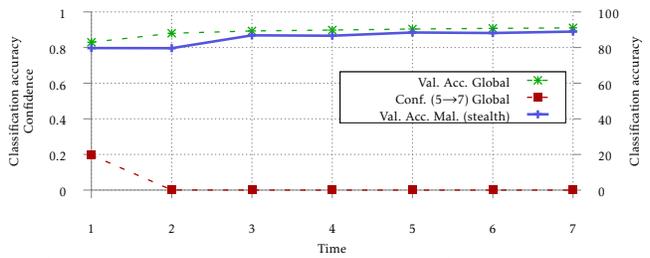}}\label{subfig: alt_min_coomed}}
	\caption{\textbf{Additional results for attacks on Byzantine-resilient aggregation mechanisms.}}
	\label{fig: krum_plots}
	\vspace{-10pt}
\end{figure}

In Section \ref{sec: byzantine_attack}, we presented the results of successful attacks on two different Byzantine resilient aggregation mechanisms: Krum \cite{peva17} and coordinate-wise median (\textsf{coomed}) \cite{yin2018byzantine}. In this section, we present the results for targeted model poisoning when Krum is used (Figure \ref{subfig: baseline_krum}). The attack uses a boosting factor of $\lambda=2$ with $k=10$. Since there is no need to overcome the constant scaling factor $\alpha_m$, the attacks can use a much smaller boosting factor $\lambda$ to ensure the global model has the targeted backdoor. With the targeted model poisoning attack, the malicious agent's update is the one chosen by Krum for 34 of 40 time steps but this causes the validation accuracy on the global model to be extremely low. Thus, our attack causes Krum to converge to an ineffective model, in contrast to its stated claims of being Byzantine-resilient. However, our attack does not achieve its goal of ensuring that the global model converges to a point with good performance on the test set due to Krum selecting just a single agent at each time step.

We also consider the effectiveness of the alternating minimization attack strategy when \textsf{coomed} is used for aggregation. While we have shown targeted model poisoning to be effective even when \textsf{coomed} is used, Figure \ref{subfig: alt_min_coomed} demonstrates that alternating minimization, which ensures that the local model learned at the malicious agent also has high validation accuracy, is not effective. 

\begin{figure}
	\centering
	\subfloat[Targeted model poisoning weight update distribution]{\includegraphics[width=0.49\textwidth]{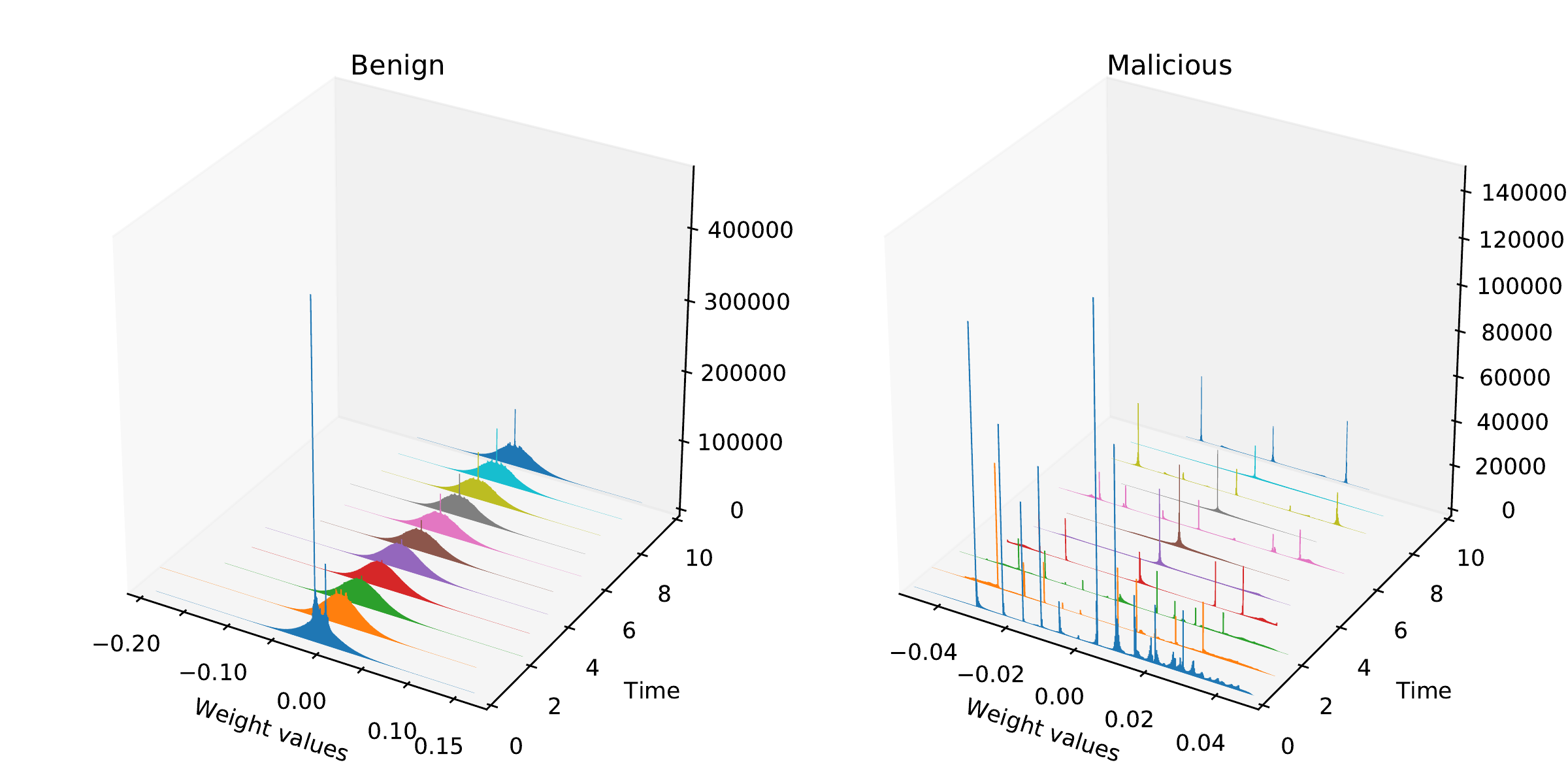}\label{subfig: converge_weights_3d}}
	\subfloat[Stealthy model poisoning weight update distribution]{\includegraphics[width=0.49\textwidth]{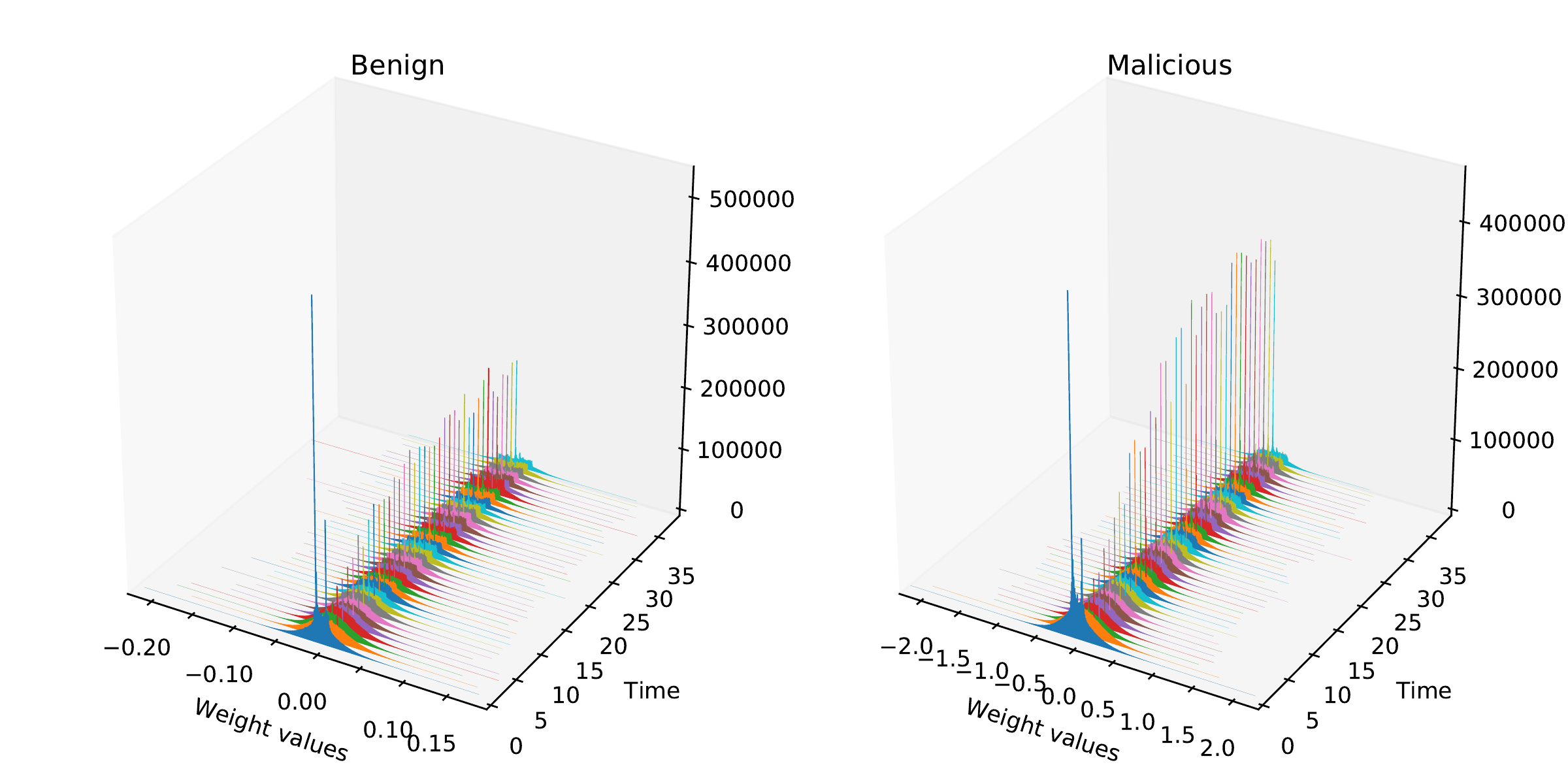}\label{subfig: converge_concat_weights_3d}}
	\hspace{0mm}
	\subfloat[Alternating minimization (only loss stealth) weight update distribution]{\includegraphics[width=0.49\textwidth]{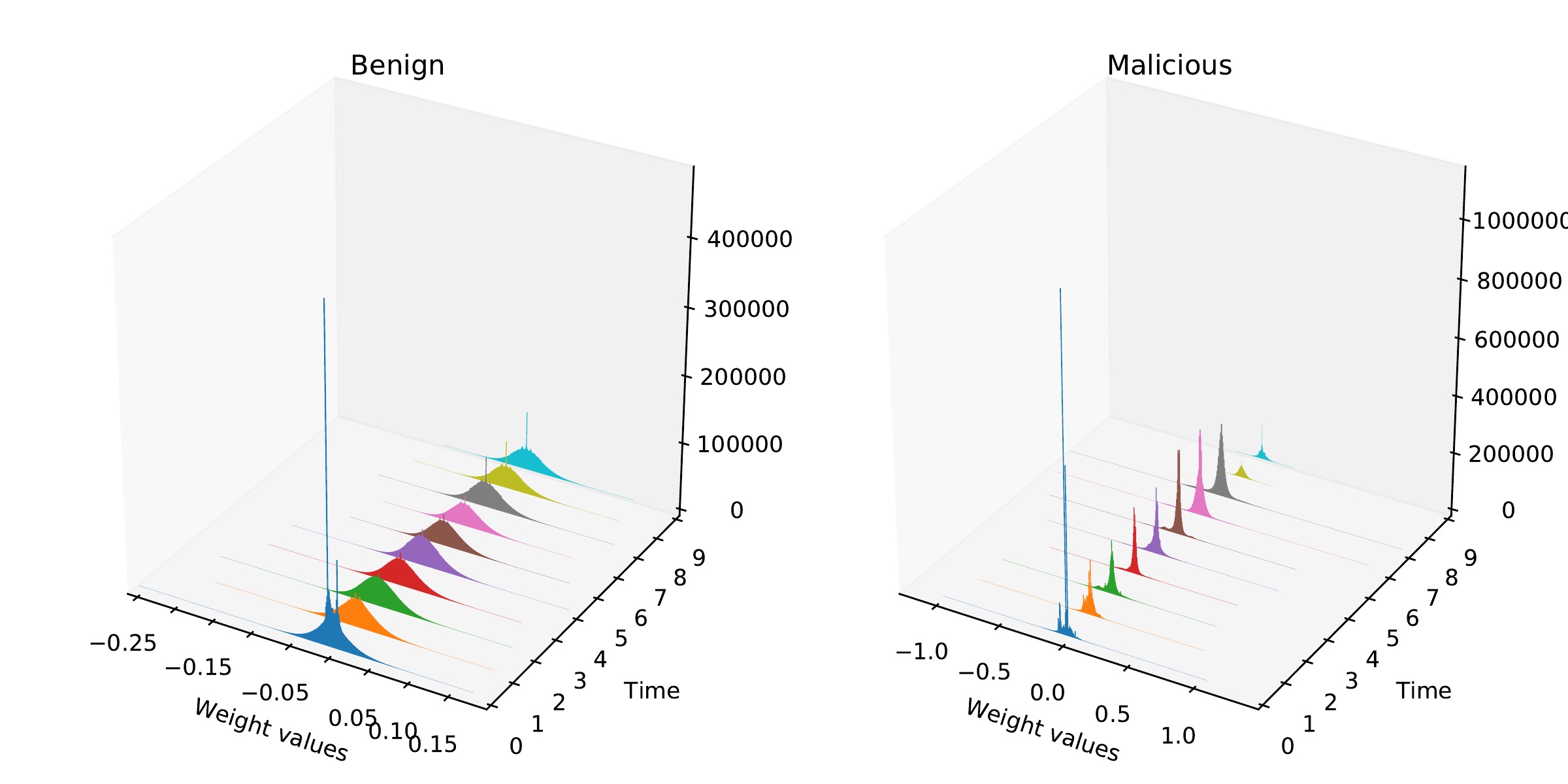}\label{subfig: alternate_weights_3d}}
	\subfloat[Alternating minimization (both loss and distance stealth) distance constraints weight update distribution]{\includegraphics[width=0.49\textwidth]{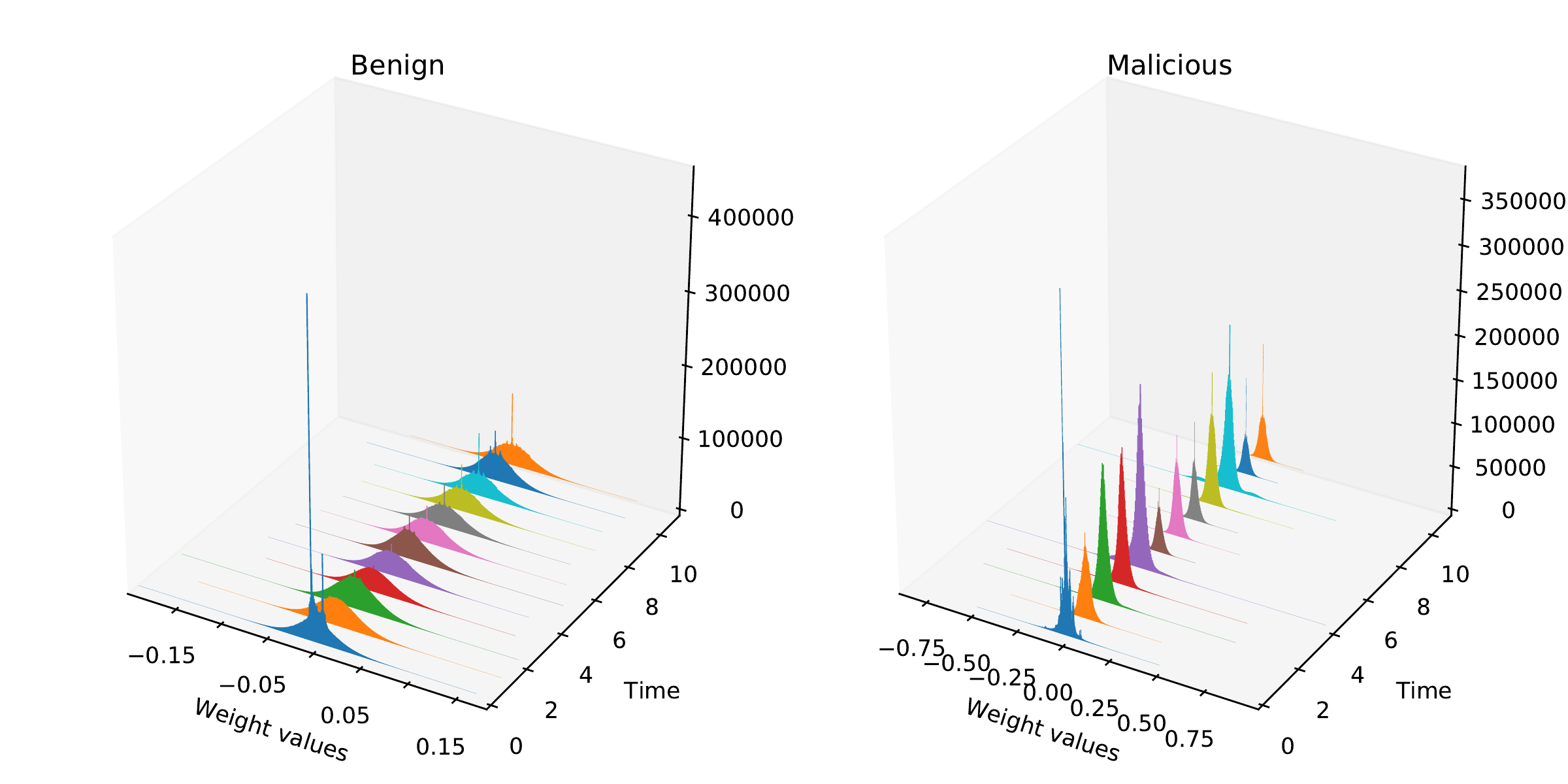}\label{subfig: alternate_dist_weights_3d}}
	\caption{Weight update distribution evolution over time for all attacks on a CNN for the Fashion MNIST dataset.}
	\label{fig: 3d_weights}
	\vspace{-10pt}
\end{figure}

\section{Visualization of weight update distributions} \label{appsec: 3d_weights}
Figure \ref{appsec: 3d_weights} shows the evolution of weight update distributions for the 4 different attack strategies on the CNN trained on the Faishon MNIST dataset. Time slices of this evolution were shown in the main text of the paper. The baseline and concatenated training attacks lead to weight update distributions that differ widely for benign and malicious agents. The alternating minimization attack without distance constraints reduces this qualitative difference somewhat but the closest weight update distributions are obtained with the alternating minimization attack with distance constraints.

\section{Interpretability for benign inputs}
\begin{figure}
	\includegraphics[width=0.95\textwidth]{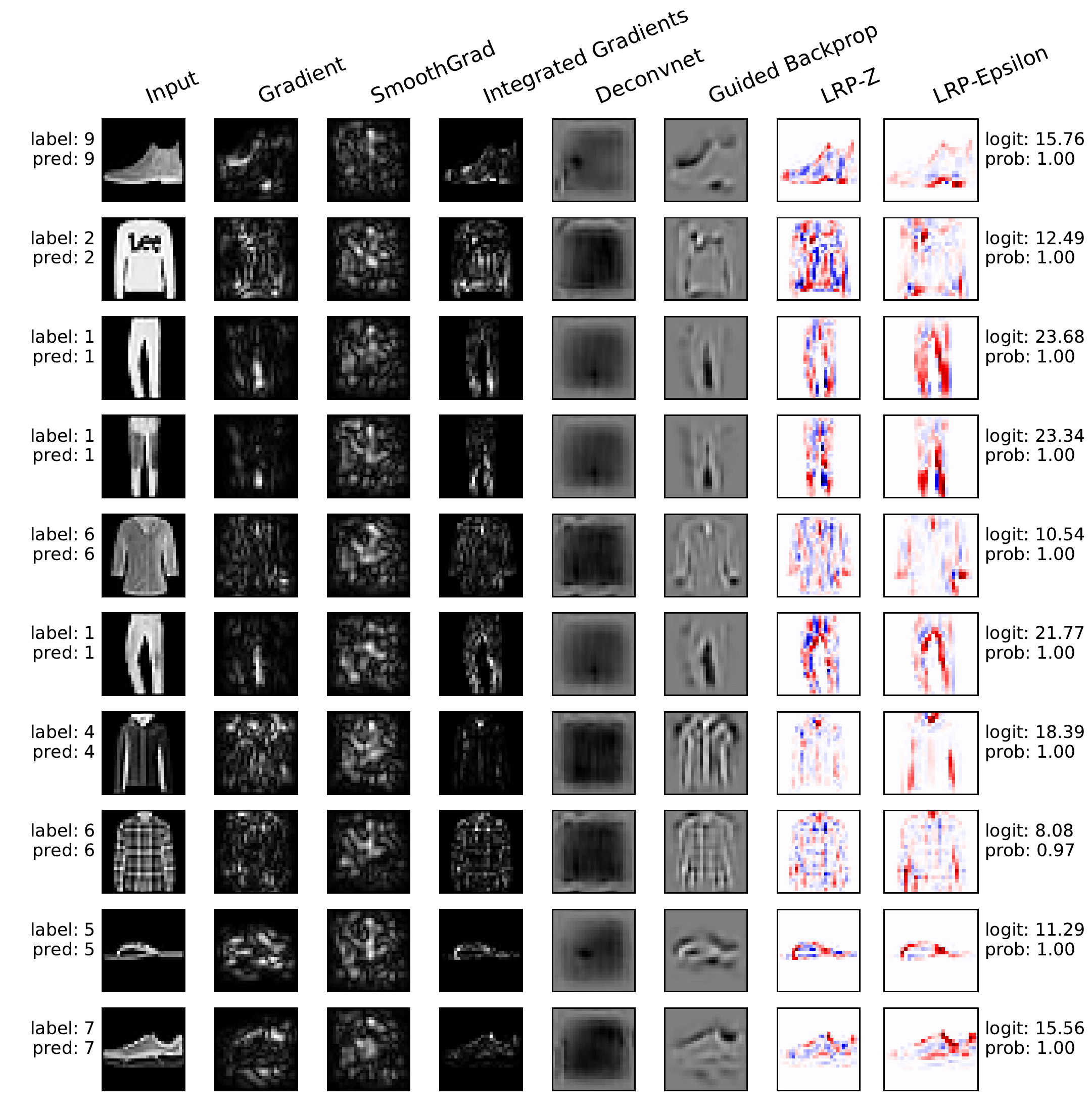}
	\caption{Decision visualizations for global model trained on Fashion MNIST data using only benign agents \emph{on benign data.}}
	\label{fig: ben_on_ben_data}
\end{figure}

\begin{figure}
	\includegraphics[width=0.95\textwidth]{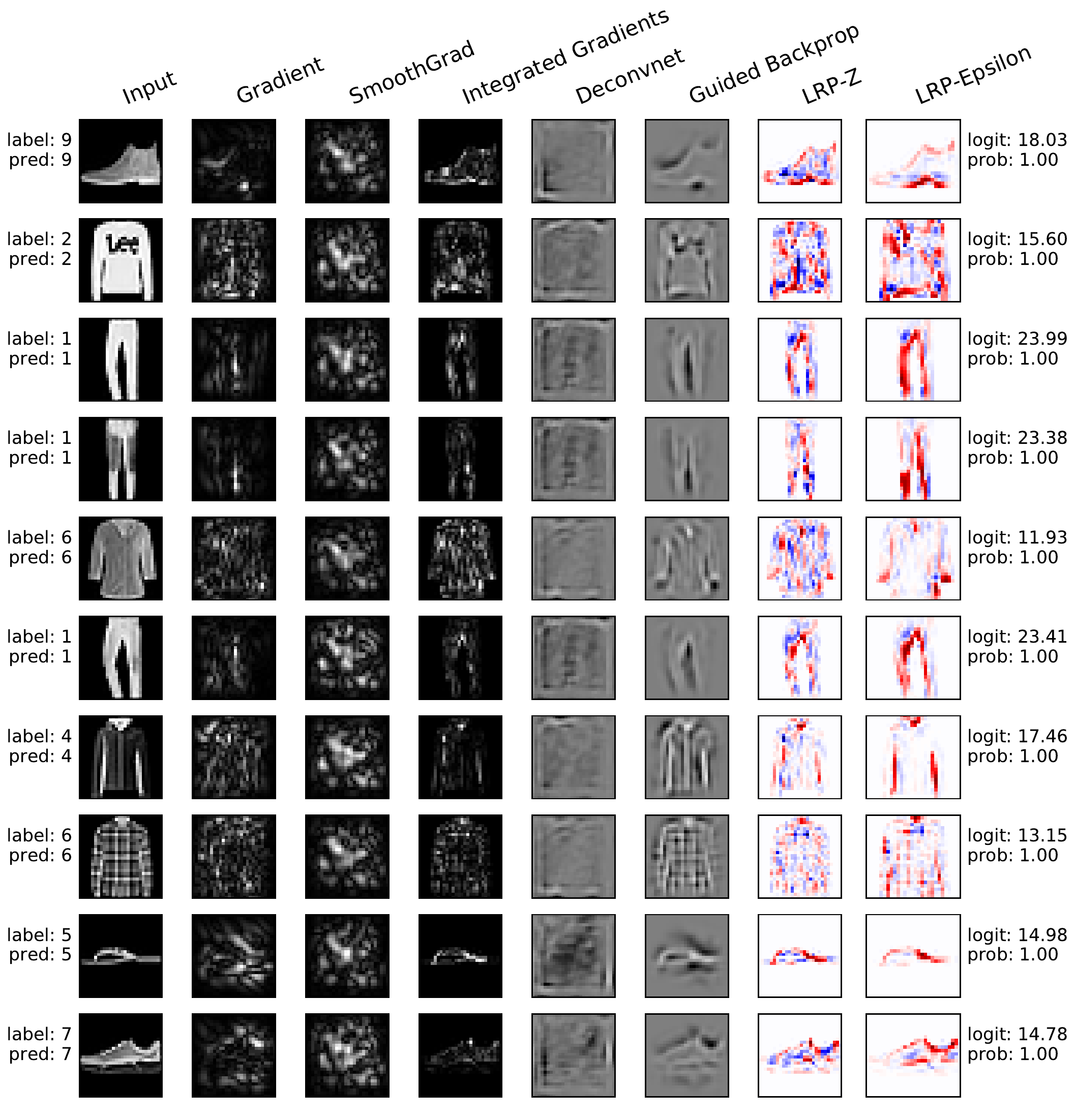}
	\caption{Decision visualizations for global model trained on Fashion MNIST data using 9 benign agents and 1 malicious agent using the baseline attack \emph{on benign data.}}
	\label{fig: mal_on_ben_data}
\end{figure}

We provide additional interpretability results for global models trained with and without the presence of a malicious agent on benign data in Figures \ref{fig: ben_on_ben_data} and \ref{fig: mal_on_ben_data} respectively. These show that the presence of the malicious agent using targeted model poisoning does not significantly affect how the global model makes decisions on benign data.

%

\end{document}